\documentclass[letterpaper, 10 pt, conference]{ieeeconf}  

\IEEEoverridecommandlockouts                              

\usepackage{graphics} 
\usepackage{bm}
\usepackage{float} 
\usepackage{stfloats}
\usepackage{caption}
\usepackage{cancel}
\usepackage{subcaption}
\usepackage{widetext}
\usepackage{epsfig} 
\usepackage{mathptmx} 
\usepackage{times} 
\usepackage{booktabs}
\usepackage{amsmath} 
\usepackage{amssymb}  
\usepackage[table, dvipsnames, svgnames, x11names]{xcolor} 
\usepackage{colortbl}
\usepackage{multirow}
\usepackage{marvosym}
\definecolor{mygray}{gray}{.9}
\usepackage{pifont}
\newcommand{\cmark}{\ding{51}}%
\newcommand{\xmark}{\ding{55}}%
\usepackage{cite}
\let\labelindent\relax
\usepackage{enumitem}
\usepackage[colorlinks,
            linkcolor=blue,
            anchorcolor=blue,
            citecolor=green,
            urlcolor=Aqua,
            backref=page]{hyperref}

\title{\LARGE \bf
A Versatile and Efficient Reinforcement Learning Framework for Autonomous Driving
}

\author{Guan Wang$^{1,4\dag}$, Haoyi Niu$^{2,4\dag}$, Desheng Zhu$^{3,4}$, Jianming Hu$^{2}$, Xianyuan Zhan$^{4}\textsuperscript{\Letter}$ and Guyue Zhou$^{4}$
\thanks{\dag Work done with equal contribution.}
\thanks{$^{1}$Department of Computer Science and Technology, Tsinghua University $^{2}$Department of Automation, Tsinghua University $^{3}$School of Mechanical Electronic and Information Engineering, China University of Mining and Technology $^{4}$Institute for AI Industry Research (AIR), Tsinghua University. Correspondence to: Xianyuan Zhan 
{\tt\small zhanxianyuan@air.tsinghua.edu.cn}}
\thanks{Please visit \url{http://carzero.xyz} for the supplementary video.}
}

\begin{document}

\maketitle
\thispagestyle{empty}
\pagestyle{empty}


\begin{abstract}
    Heated debates continue over the best autonomous driving framework. The classic modular pipeline is widely adopted in the industry owing to its great interpretability and stability, whereas the fully end-to-end paradigm has demonstrated considerable simplicity and learnability along with the rise of deep learning. As a way of marrying the advantages of both approaches, learning a semantically meaningful representation and then use in the downstream driving policy learning tasks provides a viable and attractive solution. However, several key challenges remain to be addressed, including identifying the most effective representation, alleviating the sim-to-real generalization issue as well as balancing model training cost.
    In this study, we propose a versatile and efficient reinforcement learning framework and build a fully functional autonomous vehicle for real-world validation.
    Our framework shows great generalizability to various complicated real-world scenarios and superior training efficiency against the competing baselines.

\end{abstract}

\section{Introduction}
The past decade has witnessed a surge of research interests in end-to-end autonomous driving systems\cite{Tampuu2021} driven by imitation learning (IL)\cite{Bojarski2016,Anderson2018,Muller2018,Hecker2020a,huanglearning2020,Huang2021,Wang2021} or reinforcement learning (RL)\cite{Kiran2021}. Despite its appealing simplicity and learnability endowed with  a neural network, most attempts have shown performance or adaptability issues in real-world scenarios due to large visual and dynamics gaps~\cite{rao2020,Peng2018a} between simulation and real-world environments. 
These unsolvable issues remind us of the conventional modular pipeline \cite{Levinson2011} that divides the system into modules for error-tracking and enables vehicles to behave predictably. However, this approach lacks flexibility and leads to tedious human engineering in devising complicated rules and model fine-tuning. 

The corresponding characteristics naturally inspire studies to investigate the marriage between end-to-end driving and modular pipeline, such as uncovering a series of intermediate representations as inputs for the driving decision-making model via affordance learning or image-to-image translation. Affordance learning~\cite{chen2015deepdriving, sun2019, sauer18a} predicts a heavily compressing representation for driving decision-making, e.g. comprising the distances to surrounding vehicles, heading angles, etc. The substantially low dimensionality hinders the downstream controller with brittleness and sensitivity, since little prediction error on the representation induces large deviation on the decision output. The image-to-image translation is to transform the simulated images into photo-realistic ones by domain adaptation techniques like CycleGAN~\cite{Bewley2019}. Although this alleviates the sim-to-real generalization issues in IL or RL-based autonomous driving approaches, it does not extract an effective representation to reduce the complexity of the downstream decision-making task.
We highlight the necessity of learning a suitable representation with both efficient features and sufficient information.
Contrary to the trend of pursuing complex and heavy-weighted solutions, we opt for a simple yet effective approach that offers competitive real-world adaptability, i.e. mapping the sensor inputs into an appropriately low-dimensional and semantic-meaningful representation\cite{huanglearning2020,Muller2018}, which can greatly ease the burden of learning complex driving policies with methods like RL~\cite{lesort2018}. 

In this paper, we present a deployed versatile and efficient autonomous driving framework, which contains two steps:
1) \textbf{Representation Learning} copes with the raw input from camera images and uses advanced scene understanding models \cite{cordts2016cityscapes} to perform drivable space and lane boundary identification, which empowers the system with versatility and adaptability since the inputs from different domains would be mapped into a canonical representation space. In parallel, the lower-dimensional post-perception outputs reduce the load and fully unleash the potential of the follow-up RL policy training; 2) \textbf{Distributed RL} learns RL-based driving policies from the scene understanding outputs via a carefully designed RL model with a highly efficient distributed acceleration training scheme, which enjoys highly efficient off-policy RL training. 
Given the actions from the learned RL policy, mature PID controllers are used to regulate and control the low-level driving commands. 
We evaluate our method against multiple baselines in CARLA simulator\cite{dosovitskiy17a} with intentionally introduced visual and dynamics gaps to demonstrate the real-world adaptability. Meanwhile, it is notable that our solution reduces the training time consumption by a fair margin. To underpin real-world validation, this study also makes a major contribution by building a real autonomous vehicle to deploy the proposed framework. In the subsequent real-world experiments, our system proves to be capable of generalizing to diverse complicated scenarios with varying road topology and lighting conditions, as well as the presence of obstacles. 




\section{Related Work}\label{2}
\subsection{Modular VS End-to-End Autonomous Driving}\label{2-1}
Currently, mainstream architectures for autonomous driving stem from either a modular or an end-to-end way. 

The conventional \textbf{modular pipeline} comprises a multitude of sub-systems, such as perception, localization, prediction, planning, and control\cite{Levinson2011,Yurtsever2020}. In order to handle as many real-world scenarios as possible, the interoperation of these highly functional modules relies on human-engineered deterministic rules.
The first in-depth discussions and analyses of the low-level software that encapsulates these rules are in the famous DARPA Challenge\cite{thrun2006,montemerlo2008}, paving the way for many later studies\cite{Levinson2011}. 
The modular pipeline is widely adopted in the industry due to its remarkable interpretability.
It endows the overall system with the ability to track down and locate sub-system malfunctions. Nevertheless, modular formalism bears several major drawbacks\cite{Tampuu2021}. Large amounts of human engineering are involved to fine-tune individual and cross-module configurations. The framework also suffers from severe compounding errors when decomposing the whole system into lots of smaller-scale but interpretable modules, as the perception uncertainty and modeling errors would snowball through the pipeline.

Alternatively, \textbf{end-to-end} autonomous driving has been increasingly acknowledged. It is widely accepted that human driving is a behavior reflex task\cite{Tampuu2021} that requires little high-level reasoning and conscious attention, which harbors the same view with the end-to-end architecture. As a deep learning-based solution, the task-specific end-to-end learning ability brings a great reduction of human engineering efforts. However, many studies\cite{Zablocki2021,Muller2018,Xiao2020} recognize the drawback of lack of interpretability in the development of end-to-end systems due to its black-box learning scheme.
To alleviate this limitation,
Chen\textit{ et al.}\cite{chen2021} combine the probabilistic graphical modeling with RL to improve the interpretability for autonomous driving in simulation scenarios.
Despite these efforts, weak interpretability remains a fatal disadvantage of the end-to-end approaches against the modular architecture\cite{Xiao2020}.

Based on the above comparisons, there is a growing consensus among researchers that combining the merits of both modular and end-to-end formalism could be a promising solution.
The raw perceptual inputs are processed using semantic segmentation methods\cite{cordts2016cityscapes}, or image-translation networks \cite{Bewley2019}. The intermediate outputs are then used for waypoints planning\cite{Muller2018} or control decision learning\cite{Behl2020} in an end-to-end manner.
This combined framework alleviates the drawbacks of both approaches and offers enhanced learnability, robustness, and even transferability. 

\subsection{Imitation Learning VS Reinforcement Learning}\label{2-2}
In end-to-end framework, \textbf{IL}\cite{Bojarski2016,Anderson2018,Muller2018,Hecker2020a,huanglearning2020,Huang2021,Wang2021} has been identified as a practical paradigm for autonomous driving. It uses a supervised learning scheme that 
imitates the human experts' demonstrations using algorithms like behavior cloning (BC) \cite{Behl2020,Prakash2020ExploringDA,pmlr-v100-chen20a}.
IL is known to have several major disadvantages \cite{Tampuu2021}.
First, since IL performs supervised learning on training datasets, when facing unseen scenarios during closed-loop testing, serious \textit{distribution shift} problems\cite{ross11a} take place and the agent has no idea what to do.
Second, \textit{data bias}\cite{Codevilla2019} deeply hurts the generalizability of IL policy, as the training process pays little attention to rare and risky scenarios, namely the long-tailed problem in self-driving\cite{mao2021}.
Finally, \textit{causal confusion}\cite{Haan2019} is another problem that occurs in IL, as it performs pure data fitting and handles spurious correlations in data badly.
Due to the above limitations, typical IL models can succeed in simple tasks like lane following but underperform in more complicated and rare traffic events.
To improve the performance in harder scenarios, conditional imitation learning (CIL)\cite{CodevillaICRA2018,Xiao2020,Hawke2020a} introduces a latent state to fully explain the data, thus resulting in a better model expressiveness. However, due to the lack of the ability to perform long-term predictive ``reasoning", it still has some safety issues during closed-loop testing\cite{Hawke2020a}.

Due to the ability to solve complex tasks as well as perform long-term optimization, \textbf{RL} has become another popular choice for investigating end-to-end autonomous driving \cite{Huang2020,Kiran2021,osinski2020simulation,Liang_2018_ECCV}.
RL learns how to map observations to optimized actions by maximizing the expected cumulative reward\cite{sutton1998introduction}, and can learn strong policies in a simulator without real-world labels \cite{Pan2017b}.
Although RL has lower data efficiency than IL, it could be easily implemented in some high-fidelity simulators, e.g. CARLA\cite{dosovitskiy17a}, where agents can explore in more diverse scenarios. This interactive learning ability resolves the distribution shift, data bias, and causal confusion issues in IL.
However, RL is also much harder to train with high-dimensional states. A framework with a dedicated representation learning can effectively ease the burden of RL, and also lead to more stable policies. 
Furthermore, due to the interactive learning nature of RL and its reliance on a simulator, considerable efforts need to be taken to properly address the sim-to-real issue in a deployable autonomous driving system.

\begin{figure*}[t]
\centering
\includegraphics[width=\textwidth, keepaspectratio]{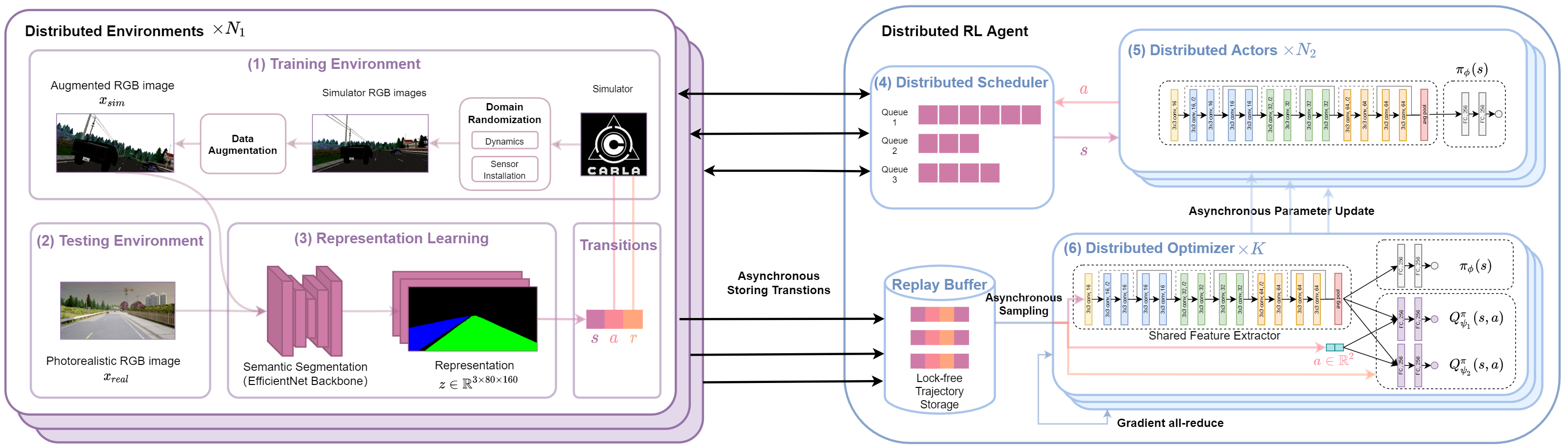}
\caption{The Overall Architecture}
\vspace{-20pt}
\label{fig_framework}
\end{figure*}

\section{Methodology}\label{3}

The overall architecture is depicted in Fig.\ref{fig_framework}, decoupling the end-to-end RL paradigm into representation learning, RL-based decision making, and an additional controller for real-world deployment: \textbf{(1) Representation Learning}: we perform drivable space and lane boundary estimation efficiently based on the input image $x$ from the monocular camera, as supervised input state-encoder.
\textbf{(2) Distributed RL-based Decision Making}: using the learned representation $z$, it learns the optimized high-level vehicle control actions $\bm{a}$, consisting of throttle $\tau$ and steering angle $\theta$, with a carefully designed fully distributed RL infrastructure, boosting data and computation efficiency. Moreover, domain generalization methods are applied to improve sim-to-real transfer performance. 
Finally, the controller maps high-level actions $\bm{a}$ to low-level vehicle control commands according to the vehicle physical properties using PID controllers.

\subsection{Representation Learning}\label{3-2}

We choose the results of drivable area segmentation and lane boundary identification \cite{bdd100k} as the intermediate representation $z$, as it is an effective way to represent road situations from monocular RGB images and is widely used in autonomous driving systems.
It labels every pixel as drivable (lane or area currently occupied by ego-vehicle), alternatively drivable (requires a lane-change), or non-drivable (blocked by obstacles) space. 

Crucially, this procedure acts as a powerful supervised state-encoder that maps camera-captured RGB images from simulation $x_{sim}$ or real world $x_{real}$ to a domain-agnostic representation $z$, effectively bridging the sim-to-real visual gap. 
Compared to unsupervised state-encoders like image translation networks adopted by\cite{Bewley2019}, semantic segmentation yields low-dimensional and human-readable intermediate representations, containing not only efficient but also sufficient information needed for RL. Notably, this disentanglement of redundant features substantially permits versatility and interpretability that facilitate downstream policy training and transfer. 

Our semantic segmentation model utilizes atrous convolution \cite{chen2017rethinking} to identify road objects with different scales, along with encoder-decoder architecture \cite{Chen2018DeepLab} to capture fine-grained details like lane lines. We introduce the light-weighted EfficientNet \cite{tan2019efficientnet} backbone network to replace the large ResNet101 backbone in previous methods \cite{ronneberger2015u,Chen2018DeepLab}, which achieves better real-time performance on the onboard computer (see Table \ref{tab:perception} for details).
Various data augmentation methods (cropping, random noise, perspective transformation, etc.) are used to further improve robustness to different environment and lighting conditions.

\subsection{Distributed Reinforcement Learning}\label{3-3}
Our decision making model is an RL agent trained in the CARLA simulator \cite{dosovitskiy17a} 
using the representation obtained through semantic segmentation as inputs. In complex RL tasks like autonomous driving, 
shrinking the extensive amount of computation time with limited resources is a crying demand. To this end, we tailor a fully distributed RL framework (illustrated in Fig.\ref{fig_framework}) for sample efficient off-policy RL training, with domain generalization and specially designed network architecture, to efficiently exploit system resources and improve final performance. 


 
 

\subsubsection{Problem formulation}

We formulate the driving decision-making problem as a Markov Decision Process (MDP) defined by a tuple $(S, A, r, T, \rho, \gamma)$, where 
$T(s'|s,a)$ denotes the transition dynamics, $\rho$ is the initial state distribution and $\gamma\in (0,1)$ is the discount factor. In our problem, $S$, $A$ and $r$ are defined as follows:

\textbf{States $S$:} 
we use the drivable area segmentation $z$ from representation learning, concatenated with the vehicle speed $v$ as the state.

\textbf{ Actions $A$:} We consider the steering angle $\theta$ and throttle $\tau$ of vehicle as actions $a = [\theta, \tau]$, where $\theta, \tau \in [-1, 1]$.

\textbf{Reward function $r$:} The design of our reward function involves concerns about four aspects: speed control $r_{speed}$, lane center keeping $r_{center}$, heading direction alignment $r_{heading}$ as well as collision and undesired lane crossing penalty $r_{penalty}$. For simplicity, we omit the state-action inputs $(s,a)$ in the reward function and describe each part as follows:
\begin{itemize}[leftmargin=*,topsep=0pt]
\item \textit{Speed control:} Instruct the vehicle with current speed $v$ to drive in the desired speed range $[v_{min}, v_{target}]$. It decays linearly when driving too slow or over-speed. $v_{max}$ indicates maximum possible speed.
\begin{align*}
r_{speed}=\min\{{v}/v_{min}, (v_{max} - v)/(v_{max} - v_{target}), 1\}
\end{align*}
\item \textit{Lane center keeping:} Instruct the vehicle to drive in the center of a lane. $d$ is the current distance from vehicle to lane center and $d_{max}$ is the maximum in-lane distance.
\begin{align*}
r_{center} = \mathrm{Clip} \left (1 - d/d_{max}, 0, 1\right )
\end{align*}
\item \textit{Heading direction:} Instruct the vehicle to drive aligned with a lane. $\alpha$ is the heading angle difference between the vehicle and the lane, with the maximum angle $\alpha_{max}$.
\begin{align*}
r_{heading} &= \mathrm{Clip}\left(1 - \alpha/\alpha_{max}, 0, 1\right)
\end{align*}
\item \textit{Collision and undesired lane crossing penalty:} Finally, we define the collision and undesired lane crossing penalty $r_{penalty}$ as follows, where $\mathbb{I}(\cdot)$ is the incident indicator:
\begin{align*}
r_{penalty} = &25 \times \mathbb{I}(\mathrm{collision}) + 12 \times \mathbb{I}(\text{cross solid line}) \\ & + 15 \times \mathbb{I}(\text{cross double solid line})
\end{align*}
\end{itemize}

The total reward at timestep $t$ is the product of $r_{speed},$ $r_{center}, r_{heading}$ minus $r_{penalty}$, enforcing a soft binary AND logic, 
expecting the agent to pursue all these goals:
\begin{align*}
r = r_{speed} \cdot r_{center} \cdot r_{heading} - r_{penalty}
\end{align*}
We solve the MDP problem using RL, which aims at learning a parameterized policy $\pi_\phi$ to maximize following expected cumulative discounted reward:
\begin{equation}
\mathbb{E}_{s_0\sim\rho, a_t\sim \pi(s),s_{t+1}\sim T(s_{t+1}|s_t,a_t)}[\sum_{t+1}^\infty \gamma^{t} r_t] \label{1}
\end{equation}

In the commonly used actor-critic RL paradigm, one optimizes the policy $\pi_\phi$ by alternatively maximizing a value function $Q_\psi(s,a)$ to approximate the cumulative return, which is learned by minimizing the squared Bellman error:
\begin{equation}
J_Q(\psi) = \mathbb{E}_{s,a,s',a'\sim \pi_\phi(s)}\left[Q_{\psi}^\pi(s,a) - (r+\gamma Q_{\psi'}^\pi(s',a')) \right]^2
\end{equation}
where $Q_{\psi'}^\pi(s,a)$ is the target Q-function, which is typically implemented as a delayed copy of the current Q-function.

\subsubsection{Distributed off-policy RL architecture}
Learning a driving policy using RL can be rather costly, we develop a fully distributed event-driven actor-critic RL formalism, which leverages multiple simulation environments, actor policies and optimizer nodes to maximally accelerate RL training.

Specifically, we use $N_1$ training environments and $N_2$ actor policies (Fig.\ref{fig_framework}{\color{blue}(1, 5)}) for parallel data collection, all of which are implemented as independent nodes. An event scheduler (Fig.\ref{fig_framework}{\color{blue}(4)}) is used to pipeline the execution of these environments and actors, making full use of computing resources.
The generated trajectories are then collected asynchronously by the replay buffer to be consumed by the off-policy RL algorithm to enhance data efficiency.

For the learning process, $K$ optimizer nodes (Fig.\ref{fig_framework}{\color{blue}(6)}) are instantiated to asynchronously update all the actor policies and Q networks.
Each optimizer node holds the same copy of all the $N_2$ policies, and also contains $M$ Q-networks (contain both current and multiple target Q-networks), leading to a total of $K\times M$ Q-networks in our framework.
The optimizers compute gradients of policies and Q-networks using batches of transitions asynchronously sampled from the replay buffer.
Distributed all-reduce is used 
to synchronize gradient shards for optimization. The final step towards completing the training loop is to asynchronously update the policies (Fig.\ref{fig_framework}{\color{blue}(5)}) with the snapshots of the policy network weights in optimizers.

\begin{figure}[t]
\centering
\includegraphics[width=\linewidth, keepaspectratio]{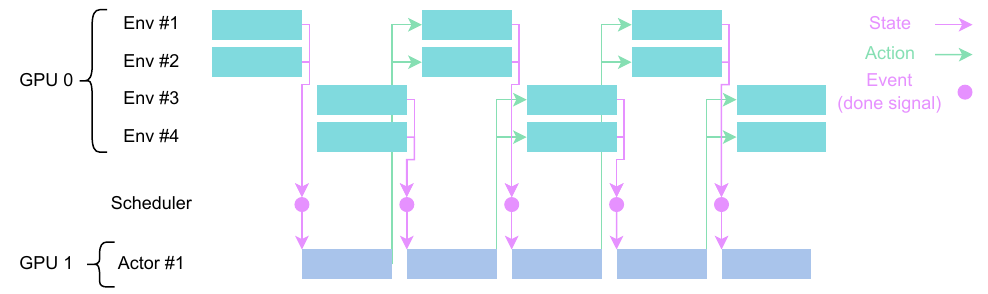}
\caption{Distributed pipeline execution}
\label{fig_pipeline}
\end{figure}

Contrary to conventional distributed architecture like IMPALA \cite{espeholt2018impala} and Ape-X \cite{horgan2018distributed}, which maintains the environment and actor in the same node, we devise a scheduler to separate all the components (environment, actor, replay buffer and optimizer), making our architecture fully distributed. 
In addition, these architectures only  access to CPU for single-step environment simulation and actor network inference.
Our distributed scheduler batches states for GPU pipelined inference (see Fig. \ref{fig_pipeline}) to minimize latency.
This also enables the flexibility to operate actors and environments on different devices (like CPU for physics simulation, or GPU for scene rendering).
Our architecture also supports centralized inference on policies \cite{espeholt2019seed},
by simply adjusting the scheduling policy.
Moreover, shared memory and lock-free data structures are used whenever possible, to minimize overhead and save bandwidth. With our distributed RL acceleration techniques, we can successfully train the whole system in less than one day on a single workstation.




\subsubsection{RL agent design}
Complex image-based reinforcement learning is prone to overfitting. A properly designed network architecture for the RL agent is a crucial part of generalization \cite{cobbe2019quantifying} and computational performance. 
We design a light-weighted network architecture for both policy and Q-networks (see Fig.\ref{fig_framework}{\color{blue}(6)}). It consists of residual convolution layers and is shared between actor policies and Q-networks to reduce the total amount of learnable parameters. 

In the optimizer node of our distributed RL architecture, we implement both the clipped double Q-learning technique in TD3 \cite{fujimoto2018addressing} to reduce the overestimation in off-policy learning, as well as the maximum entropy RL objective in soft actor-critic (SAC) \cite{haarnoja18b} to encourage exploration for improved performance. Specifically, the target Q-value in the Bellman error computation (Eq.(\ref{1})) is evaluated using multiple target Q-functions as follows:
\begin{align*}
y = r+\gamma \min_{i=1,2} Q_{\psi'_i}^\pi (s', \pi_\phi(s'))
\end{align*}
And the policy learning objective is revised to maximize both the Q-function and the entropy of the policy:
\begin{equation*}
J_\pi(\phi) = \mathbb{E}_s\mathbb{E}_{a\sim \pi_\phi}[Q_\psi^\pi(s,a)-\lambda \log \pi_\phi(s)]
\end{equation*}


\subsubsection{Sim-to-real generalization}
As the RL policy learned in a simulator needs to be deployed to unseen real-world scenarios, which induces a large domain transfer gap, we focus on improving the model generalizability by introducing a domain generalization scheme during training, incorporating domain randomization and data augmentation.

Domain randomization~\cite{tobin2017domain} is a high-level data manipulation technique with special regard to the internal physical mechanisms from the perspective of generating images. In this perspective, we randomize the configurations in the simulation environment every epoch, forcing the agent to adapt to environments with diverse properties, and also avoid overfitting on certain configurations.
At the sensor level, camera position is randomly chosen from a pre-specified range, to reflect sensor installation errors in the real world.
Vehicle physical properties (size, mass, etc.) are also randomly chosen from a pre-determined set to reduce the sensitivity to vehicle dynamics.

It has been shown in past literature~\cite{laskin2020reinforcement} that applying data augmentation in RL can greatly improved model transfer performance.
We apply data augmentation to agent's visual observations, 
such as rotating and cropping the observed images, which contributes a lot to improve the data efficiency and model generalization during RL training.





\subsection{Hardware Setup}\label{3-1}

\begin{figure}[t]
\centering
\includegraphics[width=.99\columnwidth]{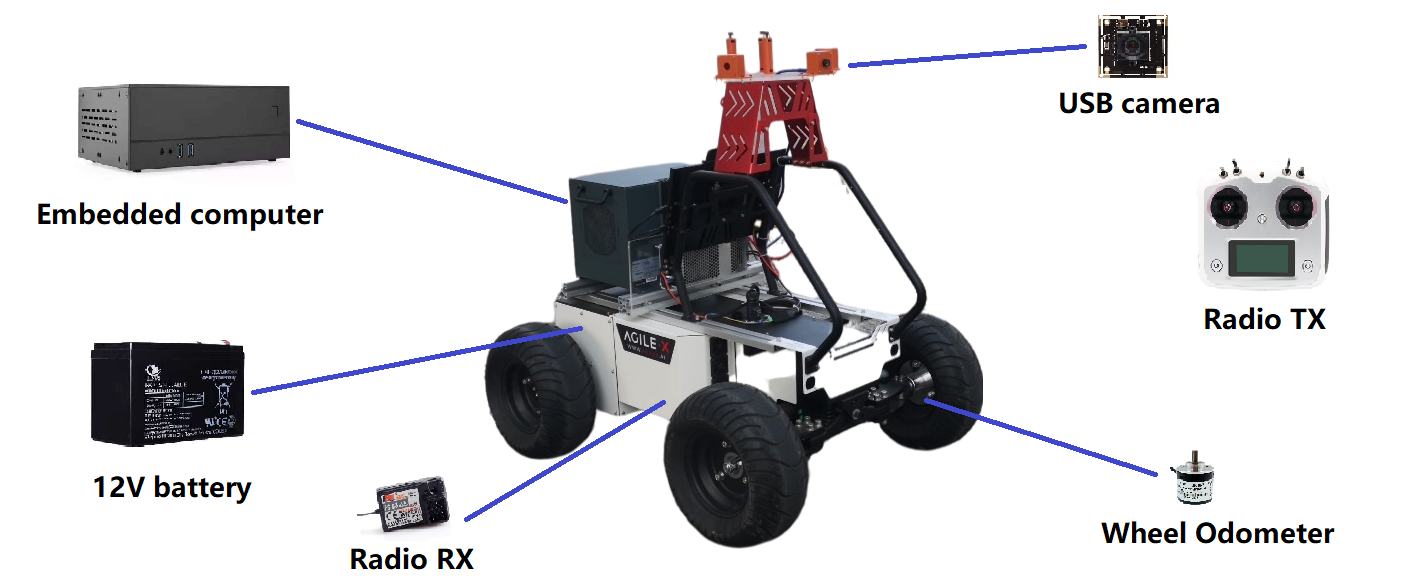}
\caption{The overall physical system}
\label{fig_hard}
\end{figure}
Our autonomous vehicle is built upon an Agile.X HUNTER Unmanned Ground Vehicle (UGV) with an onboard computer (i7-9700 CPU, 32GB RAM, GeForce RTX 3060 GPU) and a front RGB camera, as shown in Fig.\ref{fig_hard}. 
The onboard computer runs all the system modules based on Robot Operating System \cite{ros} and records trajectory statistics by Controller Area Network (CAN) in the real-world validation \ref{4-4}.
It first applies a low-pass filter on high-level control actions $\theta, \tau$ from the learned actor network to smooth out noisy signals. These stable actions are then mapped to the wheel speed $\omega_1$ and steering servo angle $\omega_2$ based on the UGV physical properties. Thereafter, these low-level control commands are delivered to UGV hardware and followed by two PID controllers. To meet the real-time requirement, we utilize quantization and computational graph optimization techniques\cite{onnxruntime} to cut down whole inference duration into 40ms latency, with 100ms control interval.


\section{Evaluation}\label{4}

In this section, we present the baselines, evaluation metrics information and detailed experiment results quantitatively.
\subsection{Experiment Settings}
\subsubsection{Baselines}

We incorporate five baselines in this study: end-to-end IL \cite{Bojarski2016}, end-to-end RL \cite{kendall2019}, CycleGAN + IL\cite{Bewley2019}, CycleGAN + RL and the IL variant of our framework (replace the RL policy to the behavior cloning policy). We re-implement the IL and RL modules of these baselines in our framework for a thorough comparison. 

For the training of all IL-based baselines, we follow the treatment described in~\cite{Muller2018}. We collect 28h of driving data using a privileged modular pipeline as expert (has access to ground-truth map and obstacle information) provided by CARLA \cite{dosovitskiy17a}, and add noise to 20\% of the expert control outputs to improve the robustness of the learned policy\cite{CodevillaICRA2018}.
For RL-based baselines, all environment and agent configurations are set the same as our proposed approach.

\subsubsection{Evaluation Metrics}
For performance evaluation, we use the following metrics:  
\begin{itemize}[leftmargin=*,topsep=0pt]
    \item \textbf{MPI (m)}: meters per intervention, indicating the level of autonomy of the vehicle, which is widely adopted in real-world autonomous vehicle testing. Interventions are performed when collisions happen or the vehicle doesn't move in more than one minute.
    \item \textbf{SR (\%)}: success rate, referring to the proportion of travelled distance from the start to the first intervention with respect to 
    the whole journey in one trial. 
    \item \textbf{Std[$\theta$] (\bm{$^\circ$})}: the standard deviation of the steering angle, reflecting the lateral smoothness of the trajectory.
    \item \textbf{Std[$v$] (m/s)}: the standard deviation of the velocity, revealing the longitudinal smoothness of the trajectory.
\end{itemize}


\subsection{Evaluation on the Representation Learning Model}
The representation model in \ref{3-2} achieves superior accuracy (mIoU) over many popular segmentation models on the BDD100k dataset \cite{bdd100k}, as shown in the results of Table \ref{Table1}. Our model also enjoys greater inference efficiency for more reliable real-time response, capable of running at 31 frames per second on the onboard computer.
\begingroup
  \setlength\tabcolsep{1.5pt}
  \begin{table}[h]
  \centering
  \caption{Perception Model Comparison}\label{tab:perception}
      \begin{tabular}{cccc}
      \toprule
      Method & Backbone & mIoU (validation)/\%  & Inference time (ms) \\ \midrule
      UNet \cite{ronneberger2015u} & ResNet101 & 87.2 & 64 \\
      DeepLabv3+ \cite{Chen2018DeepLab} & ResNet101 & 92.5 & 50 \\
      \textbf{Ours} & \textbf{EfficientNet-B0} & \textbf{93.4} & \textbf{32} \\ 
      \bottomrule
      \end{tabular}
      \label{Table1}
  \end{table}
\endgroup

\subsection{Comparison in Simulation}\label{4-3}
\begin{figure}
	\centering
	\begin{subfigure}{0.235\textwidth}
		\centering
        \includegraphics[width=\textwidth]{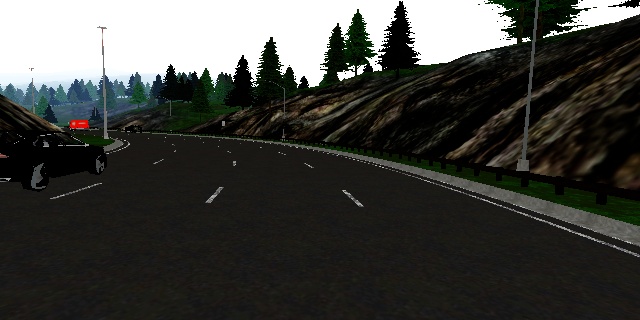}
        \caption{Training}
	\end{subfigure}
    \begin{subfigure}{0.235\textwidth}
		\centering
        \includegraphics[width=\textwidth]{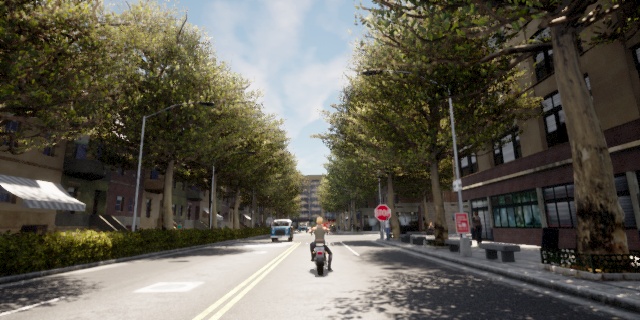}
        \caption{Testing}
	\end{subfigure}
    \caption{Examples from the \textit{NoGap} benchmark. Training scenarios use game-like low-quality coarse rendering, while testing scenarios are photo-realistic.}
    \label{fig:nogap_rgb}
\end{figure}

\begin{table*}[t]
\centering
\caption{Validation on training and testing performance in simulation, metrics are averaged over 50 runs.}
\begin{tabular}{ccccc}
\toprule
\multirow{2}{*}{Framework} & \multirow{2}{*}{Representation} & \multirow{2}{*}{Decision Making} & \multicolumn{2}{c}{MPI (m)} \\\cmidrule(r){4-5}
                    &   &                                   & Train & Test \\ \midrule
Fully End-to-end IL~\cite{Bojarski2016} & RGB & IL & 134.6 & 16.2 \\
Fully End-to-end RL~\cite{kendall2019} & RGB & RL & 24.9 & 0.4  \\
\midrule
End-to-end IL + CycleGAN~\cite{Bewley2019} & Real-to-sim Translation & IL  & 134.6 & 46.7 \\
End-to-end RL + CycleGAN~\cite{Bewley2019}& Real-to-sim Translation & RL & 24.9 & 5.3  \\\midrule
\textbf{Ours} (IL) & Semantic Representation & IL & 306.6 & 180.9 \\ 
\textbf{Ours} & Semantic Representation & RL & \textbf{449.4}  & \textbf{332.6}\\ \bottomrule
\end{tabular}
\label{tab_sim}
\end{table*}

The main challenge of deploying a simulator-trained system to real world is the
visual, vehicle dynamics and scenario gap.
Previous simulation-based autonomous driving benchmarks, such as \textit{NoCrash} \cite{Codevilla2019} and CARLA benchmark \cite{dosovitskiy17a}, do not reflect such challenges effectively. For example, the training and test scenarios are visually similar, and not suited well for comprehensive real-world deployment validation.

We propose a new \textit{NoGap} benchmark to measure the sim-to-real gaps explicitly and provide better real-world generalization evaluations.
In \textit{NoGap} benchmark. the autonomous driving system is trained in ``simulator" and then tested in ``real-world" (simulator with different configurations). The evaluation settings are described as follows:


\begin{itemize}[leftmargin=*,topsep=0pt]
\item \textbf{Visual Gap}: We use different simulator rendering modes to create intentionally introduced visual gaps. During training, low-quality coarse rendering is used, producing game-like images. While in testing, the simulator is switched to photo-realistic rendering, producing images that look like real-world. Please refer to Fig. \ref{fig:nogap_rgb} for examples.
\item \textbf{Dynamics Gap}: Vehicle physical properties are randomly chosen at the beginning of every testing episode, to validate generalization between different vehicle dynamics.
\item \textbf{Scenario Gap}: 7 CARLA maps are used for training, and 1 reserved for testing, to reflect scenario difference. 
\end{itemize}

We use the MPI metric in simulation evaluations. 
Inspired by \cite{Codevilla2019}, the environment is reset to move the vehicle to a safe state after an intervention
for precise intervention counting. Moreover, background vehicles controlled by built-in modular pipeline in CARLA are generated to emulate real-world traffic and dynamic obstacles.

As shown in Table \ref{tab_sim}, fully end-to-end frameworks fail to generalize from training to testing scenarios due to large visual gaps in RGB inputs. Image-to-image translation method (CycleGAN) can improve testing performance effectively. Furthermore, baselines of training RL with RGB images manifest poorer performance against their IL counterparts by a wide margin, due to the high-dimensional input, while the situation comes to the opposite when training on our segmentation outputs. 
Thanks to the efficient yet informative representation that eases the burden of RL, our framework outperforms all the competing baselines with high generalizability to different maps and manufactured visual gaps. 







\begin{table*}[t]
\centering
\caption{Real world evaluation ($>$ means no intervention over the whole trajectory with specified total length.)}
\begin{tabular}{c|c|c||>{\columncolor{gray!40}}c>{\columncolor{gray!30}}c>{\columncolor{gray!20}}c>{\columncolor{gray!10}}c|>{\columncolor{gray!40}}c>{\columncolor{gray!30}}c>{\columncolor{gray!20}}c>{\columncolor{gray!10}}c}
\toprule

\rowcolor{white}
 \multicolumn{3}{c||}{\multirow{2}*{\textbf{Real World Task}}} & \multicolumn{8}{c}{\multirow{2}*{\textbf{Framework}}} \\
 \multicolumn{3}{c||}{}  & \multicolumn{8}{c}{}\\
 
 \rowcolor{White}
Road & Obstacle & Lighting & \multicolumn{4}{c|}{\textbf{Ours} (IL)  } & \multicolumn{4}{c}{\textbf{Ours} } \\

Topology  & Setup  & Condition  & MPI ($m$) & SR ($\%$) & Std[$\theta$] ($^\circ$) & Std[$v$] ($m/s$) & MPI ($m$) & SR ($\%$) & Std[$\theta$] ($^\circ$) & Std[$v$] ($m/s$) \\ \midrule\midrule
         
\multirow{3}*{Straight} & \xmark & Day & 92.1 & 48.9 & \textbf{1.30} & 0.30 & $\bm{>}$\textbf{1163.5} & \textbf{100.0} & 1.72 & \textbf{0.19}\\

 & \xmark & Night & 187.1 & 49.0 & \textbf{1.67} & \textbf{0.18} & $\bm{>}$\textbf{1304.5} & \textbf{100.0} & 1.89 & 0.20 \\  
 
 & \cmark & Day & 4.1 & 16.7 & \textbf{1.25} & 0.37 & \textbf{34.5} & \textbf{75.0} & 2.59 & \textbf{0.30} \\ \midrule
 
Turn     & \xmark & Day & 7.2 & 53.2 & \textbf{3.03} & 0.32 & $\bm{>}$\textbf{214.9} & \textbf{100.0} & 3.81 & \textbf{0.23}\\ \bottomrule
\end{tabular}
\label{Table4}
\end{table*}

\subsection{Real World Benchmark}\label{4-4}

\begin{figure}
\centering
\includegraphics[width=0.48\textwidth, keepaspectratio]{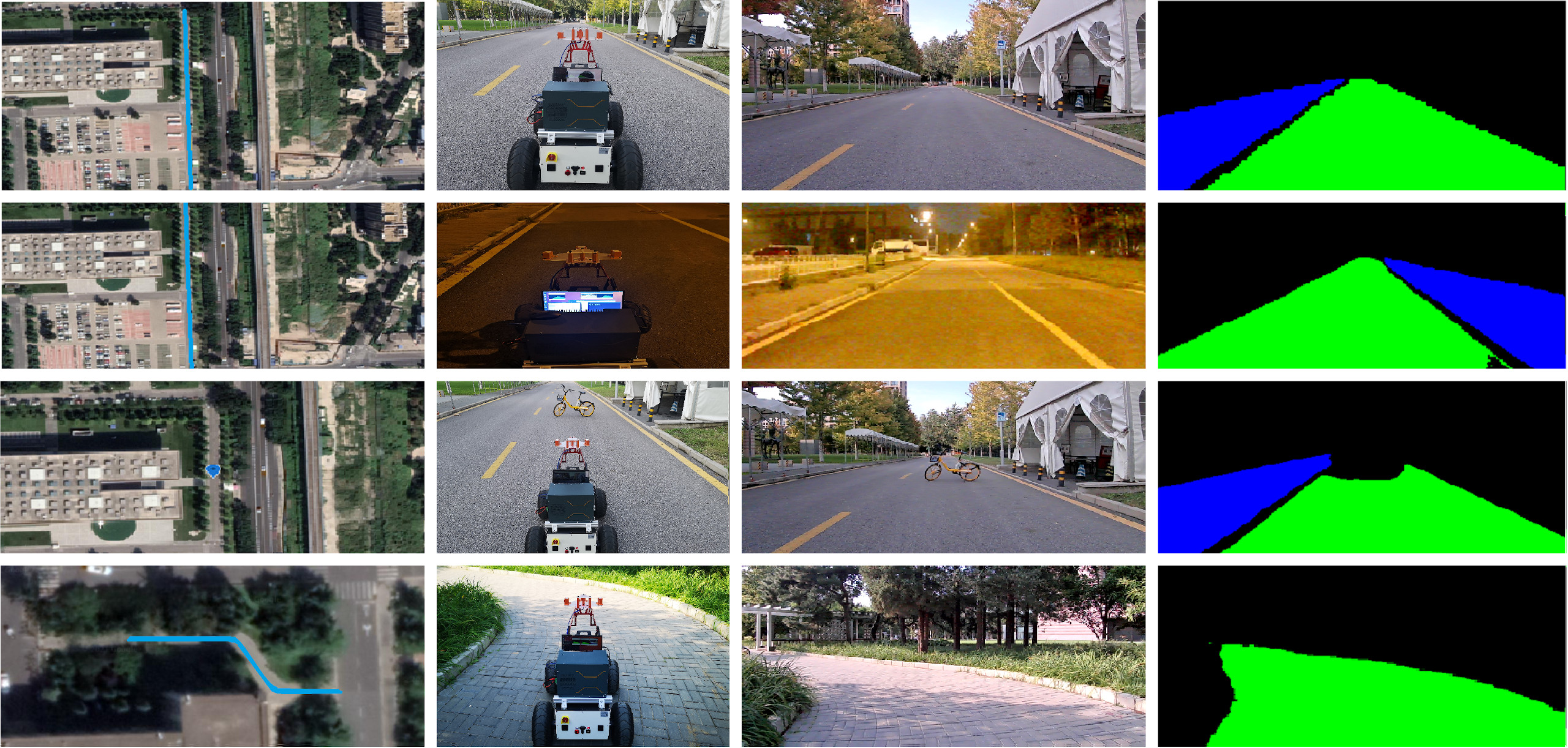}
\caption{The bird's eye views, actual scenarios, raw inputs, and semantic segmentation results (from left to right) for the tasks in Table \ref{Table4} respectively}
\label{fig_4x4}
\vspace{-5pt}
\end{figure}

\begin{figure}[h!]
    \centering
    \begin{subfigure}{.5\textwidth}
        \centering
        \includegraphics[width=\textwidth]{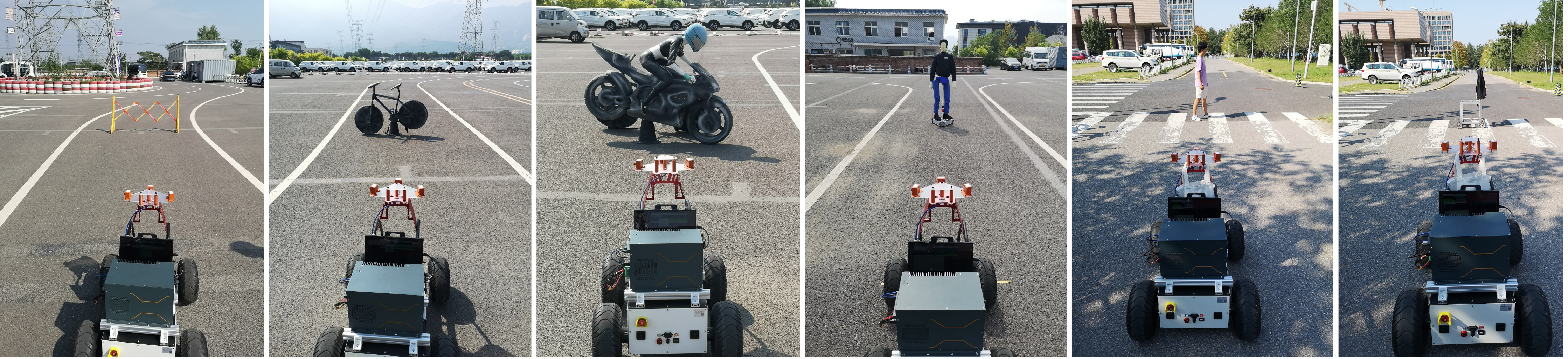}
        \caption{Obstacle avoidance tests: barrier, bicycle, motorcycle, dummy, person, and chair-like robot (from left to right)}
        \label{fig:additional_obstacle}
    \end{subfigure}
    \par\bigskip
    \vspace{-8pt}
    \begin{subfigure}{.5\textwidth}
        \centering
        \includegraphics[width=\textwidth]{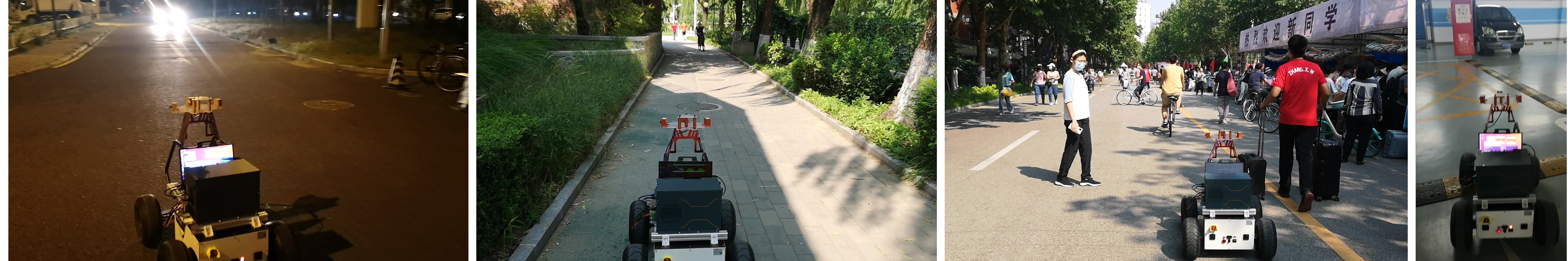}
        \caption{Test in unseen and complex scenarios: high beam, sidewalk, crowds, and garage scenarios (from left to right)}
        \label{fig:additional_scenario}
    \end{subfigure}
    \caption{Qualitative real-world experiments}
    \label{fig_realworld_additional}
    \vspace{-20pt}
\end{figure}

To evaluate the autonomous driving framework on our test vehicle described in Section \ref{3-1}, we categorize the real world tasks based on different attributes: road topology (straight and turn), obstacles setup (with obstacle ahead), and lighting condition (day-time and night-time). 
We test our framework with the most competitive baseline standing out from simulation evaluation in Section \ref{4-3}, the IL version of ours,
under the same set of tasks.

To assess the lane-keeping ability on straight and turning roads, we run the vehicle five trials per framework for each non-obstacle task, during each of which we set the vehicle to a collision-free position as close as possible to the place where the intervention takes place. The obstacle setup is implemented as the vehicle accelerates from 0m/s and tries to circumvent the obstacle 5m ahead. If the vehicle successfully bypasses the obstacle, we record it as a successful trial and reset the vehicle to the initial location for the next trial. Otherwise, it commits a failed trial and counts one more intervention.
Table \ref{Table4} presents the comparisons of real world generalizability between the IL and RL version of
our framework on tasks with different attribute settings.

Several observations can be drawn from these experimental results.
Our framework achieves much higher MPI and SR on every task, superior to IL version in terms of autonomy and safety. Ours is also generalizable to different lighting conditions, whereas surprisingly, IL version at night-time visibly outperforms itself at day-time. Ours also has smaller velocity std in most of the tasks, while the more conservative IL version has more hold-backs during the real-world tests. Finally, it is observed that ours has better ability to bypass the obstacles. This results in larger standard deviation of steering angle of ours than the more conservative IL version, as IL version directly bumps into the obstacle and scrapes the curb at turns more often.

We also perform qualitative real-world evaluation on complex and unseen scenarios (setup as Fig.\ref{fig:additional_obstacle} and \ref{fig:additional_scenario}), such as being exposed to head-on high beams, traveling on sidewalks, through dense crowds, and even in a garage. Our framework can handle lane-following, turning and dynamic obstacle avoidance smoothly, revealing good generalization performance. Please refer to supplementary video for details.




\subsection{Computation performance}
Table \ref{Table_extra} reports the computational performance of our distributed RL architecture. Due to limited computational resources (a single workstation with 2 RTX3090 GPUs), we use $K = 1$ for all experiments.
Compared to non-distributed version which completes the training process in 87.2 hours, our distributed framework ends up with only 12.6 hours, yielding a salient training speedup. With our flexible scheduling scheme, scaling up number of environments $N_1$ and actors $N_2$ also yields a consistent speed up. 
Our perception model for representation learning is also more light-weighted and effective than CycleGAN, which takes 1/5 computation time to train and also achieves better overall performance.
\begingroup
  \setlength\tabcolsep{1.5pt}
  \begin{table}[h]
  \centering
  \caption{Ablations on Training Efficiency (Unit: hours)}
      \begin{tabular}{c|c|c|c}
      \toprule
      Framework & Representation & RL @2.5M steps & Total \\ \midrule
      Non-distributed version & 5.1 & 87.2 & 92.3 \\
      CycleGAN, $N_1=4, N_2=2$ & 26.9 & 12.0 & 38.9 \\\midrule
      Ours, \textbf{$N_1=4, N_2=2$} & 5.1 & 12.6 & \textbf{17.7} \\ 
      Ours, \textbf{$N_1=4, N_2=1$} & 5.1 & 13.6 & 18.7 \\ 
      Ours, \textbf{$N_1=2, N_2=1$} & 5.1 & 18.7 & 23.8 \\ 
      \bottomrule
      \end{tabular}
      \label{Table_extra}
  \end{table}
\endgroup

\section{Conclusion}\label{5}

In this paper, we propose a versatile and efficient RL framework for autonomous driving.
By decoupling semantic-meaningful representation learning from RL, we alleviate the challenging sim-to-real gap, enhance perception performance, and improve RL decision making abilities.
We also tailor a event-driven fully distributed framework for off-policy RL, making it possible to train the whole system in less than one day on a single workstation.
We validate the performance and generalization of our framework in simulation with intentionally introduced visual and dynamics gaps.
Finally, we build an autonomous vehicle to deploy our framework for real-world evaluation.
Furthermore, our framework could generalize to diverse and unseen complex real-world scenarios. It also achieves superior performance compared with various IL and RL baselines.






\bibliographystyle{IEEEtran}
\bibliography{mylib}

\begin{thebibliography}{10}
\providecommand{\url}[1]{#1}
\csname url@rmstyle\endcsname
\providecommand{\newblock}{\relax}
\providecommand{\bibinfo}[2]{#2}
\providecommand\BIBentrySTDinterwordspacing{\spaceskip=0pt\relax}
\providecommand\BIBentryALTinterwordstretchfactor{4}
\providecommand\BIBentryALTinterwordspacing{\spaceskip=\fontdimen2\font plus
\BIBentryALTinterwordstretchfactor\fontdimen3\font minus
  \fontdimen4\font\relax}
\providecommand\BIBforeignlanguage[2]{{%
\expandafter\ifx\csname l@#1\endcsname\relax
\typeout{** WARNING: IEEEtran.bst: No hyphenation pattern has been}%
\typeout{** loaded for the language `#1'. Using the pattern for}%
\typeout{** the default language instead.}%
\else
\language=\csname l@#1\endcsname
\fi
#2}}

\bibitem{Tampuu2021}
A.~Tampuu, T.~Matiisen, M.~Semikin, D.~Fishman, and N.~Muhammad, ``A survey of
  end-to-end driving: Architectures and training methods,'' \emph{IEEE
  Transactions on Neural Networks and Learning Systems}, pp. 1--21, 2020.

\bibitem{Bojarski2016}
M.~Bojarski, D.~Testa, D.~Dworakowski, B.~Firner, B.~Flepp, P.~Goyal,
  L.~Jackel, M.~Monfort, U.~Muller, J.~Zhang, X.~Zhang, J.~Zhao, and K.~Zieba,
  ``End to end learning for self-driving cars,'' \emph{ArXiv}, vol.
  abs/1604.07316, 2016.

\bibitem{Anderson2018}
\BIBentryALTinterwordspacing
P.~Anderson, A.~Chang, D.~S. Chaplot, A.~Dosovitskiy, S.~Gupta, V.~Koltun,
  J.~Kosecka, J.~Malik, R.~Mottaghi, M.~Savva, and A.~R. Zamir, ``{On
  Evaluation of Embodied Navigation Agents},'' pp. 1--11, 2018. [Online].
  Available: \url{http://arxiv.org/abs/1807.06757}
\BIBentrySTDinterwordspacing

\bibitem{Muller2018}
\BIBentryALTinterwordspacing
M.~M{\"{u}}ller, A.~Dosovitskiy, B.~Ghanem, and V.~Koltun, ``{Driving Policy
  Transfer via Modularity and Abstraction},'' no. CoRL, 2018. [Online].
  Available: \url{http://arxiv.org/abs/1804.09364}
\BIBentrySTDinterwordspacing

\bibitem{Hecker2020a}
S.~Hecker, D.~Dai, A.~Liniger, M.~Hahner, and L.~{Van Gool}, ``{Learning
  accurate and human-like driving using semantic maps and attention},''
  \emph{IEEE International Conference on Intelligent Robots and Systems}, pp.
  2346--2353, 2020.

\bibitem{huanglearning2020}
J.~Huang, S.~Xie, J.~Sun, Q.~Ma, C.~Liu, D.~Lin, and B.~Zhou, ``Learning a
  decision module by imitating driver’s control behaviors,'' in
  \emph{Proceedings of the Conference on Robot Learning (CoRL) 2020}.

\bibitem{Huang2021}
Z.~Huang, C.~Lv, Y.~Xing, and J.~Wu, ``{Multi-Modal Sensor Fusion-Based Deep
  Neural Network for End-to-End Autonomous Driving with Scene Understanding},''
  \emph{IEEE Sensors Journal}, vol.~21, no.~10, pp. 11\,781--11\,790, 2021.

\bibitem{Wang2021}
Y.~Wang, D.~Zhang, J.~Wang, Z.~Chen, Y.~Li, Y.~Wang, and R.~Xiong, ``{Imitation
  Learning of Hierarchical Driving Model: From Continuous Intention to
  Continuous Trajectory},'' \emph{IEEE Robotics and Automation Letters},
  vol.~6, no.~2, pp. 2477--2484, 2021.

\bibitem{Kiran2021}
B.~R. Kiran, I.~Sobh, V.~Talpaert, P.~Mannion, A.~A. Sallab, S.~Yogamani, and
  P.~Perez, ``{Deep Reinforcement Learning for Autonomous Driving: A Survey},''
  \emph{IEEE Transactions on Intelligent Transportation Systems}, no. February,
  2021.

\bibitem{rao2020}
K.~Rao, C.~Harris, A.~Irpan, S.~Levine, J.~Ibarz, and M.~Khansari,
  ``Rl-cyclegan: Reinforcement learning aware simulation-to-real,'' in
  \emph{2020 IEEE/CVF Conference on Computer Vision and Pattern Recognition
  (CVPR)}, 2020, pp. 11\,154--11\,163.

\bibitem{Peng2018a}
X.~B. Peng, M.~Andrychowicz, W.~Zaremba, and P.~Abbeel, ``{Sim-to-Real Transfer
  of Robotic Control with Dynamics Randomization},'' \emph{Proceedings - IEEE
  International Conference on Robotics and Automation}, pp. 3803--3810, 2018.

\bibitem{Levinson2011}
J.~Levinson, J.~Askeland, J.~Becker, J.~Dolson, D.~Held, S.~Kammel, J.~Z.
  Kolter, D.~Langer, O.~Pink, V.~Pratt, M.~Sokolsky, G.~Stanek, D.~Stavens,
  A.~Teichman, M.~Werling, and S.~Thrun, ``Towards fully autonomous driving:
  Systems and algorithms,'' in \emph{2011 IEEE Intelligent Vehicles Symposium
  (IV)}, 2011, pp. 163--168.

\bibitem{chen2015deepdriving}
C.~Chen, A.~Seff, A.~Kornhauser, and J.~Xiao, ``Deepdriving: Learning
  affordance for direct perception in autonomous driving,'' in
  \emph{Proceedings of the IEEE international conference on computer vision},
  2015, pp. 2722--2730.

\bibitem{sun2019}
C.~Sun, J.~M.~U. Vianney, and D.~Cao, ``Affordance learning in direct
  perception for autonomous driving,'' \emph{arXiv preprint arXiv:1903.08746},
  2019.

\bibitem{sauer18a}
\BIBentryALTinterwordspacing
A.~Sauer, N.~Savinov, and A.~Geiger, ``Conditional affordance learning for
  driving in urban environments,'' in \emph{Proceedings of The 2nd Conference
  on Robot Learning}, ser. Proceedings of Machine Learning Research,
  A.~Billard, A.~Dragan, J.~Peters, and J.~Morimoto, Eds., vol.~87.\hskip 1em
  plus 0.5em minus 0.4em\relax PMLR, 29--31 Oct 2018, pp. 237--252. [Online].
  Available: \url{https://proceedings.mlr.press/v87/sauer18a.html}
\BIBentrySTDinterwordspacing

\bibitem{Bewley2019}
A.~Bewley, J.~Rigley, Y.~Liu, J.~Hawke, R.~Shen, V.~D. Lam, and A.~Kendall,
  ``{Learning to drive from simulation without real world labels},''
  \emph{Proceedings - IEEE International Conference on Robotics and
  Automation}, vol. 2019-May, pp. 4818--4824, 2019.

\bibitem{lesort2018}
T.~Lesort, N.~D{\'\i}az-Rodr{\'\i}guez, J.-F. Goudou, and D.~Filliat, ``State
  representation learning for control: An overview,'' \emph{Neural Networks},
  vol. 108, pp. 379--392, 2018.

\bibitem{cordts2016cityscapes}
M.~Cordts, M.~Omran, S.~Ramos, T.~Rehfeld, M.~Enzweiler, R.~Benenson,
  U.~Franke, S.~Roth, and B.~Schiele, ``The cityscapes dataset for semantic
  urban scene understanding,'' in \emph{Proc. of the IEEE Conference on
  Computer Vision and Pattern Recognition (CVPR)}, 2016.

\bibitem{dosovitskiy17a}
\BIBentryALTinterwordspacing
A.~Dosovitskiy, G.~Ros, F.~Codevilla, A.~Lopez, and V.~Koltun, ``{CARLA}: {An}
  open urban driving simulator,'' in \emph{Proceedings of the 1st Annual
  Conference on Robot Learning}, ser. Proceedings of Machine Learning Research,
  S.~Levine, V.~Vanhoucke, and K.~Goldberg, Eds., vol.~78.\hskip 1em plus 0.5em
  minus 0.4em\relax PMLR, 13--15 Nov 2017, pp. 1--16. [Online]. Available:
  \url{http://proceedings.mlr.press/v78/dosovitskiy17a.html}
\BIBentrySTDinterwordspacing

\bibitem{Yurtsever2020}
E.~Yurtsever, J.~Lambert, A.~Carballo, and K.~Takeda, ``{A Survey of Autonomous
  Driving: Common Practices and Emerging Technologies},'' \emph{IEEE Access},
  vol.~8, pp. 58\,443--58\,469, 2020.

\bibitem{thrun2006}
\BIBentryALTinterwordspacing
S.~Thrun, M.~Montemerlo, H.~Dahlkamp, D.~Stavens, A.~Aron, J.~Diebel, P.~Fong,
  J.~Gale, M.~Halpenny, G.~Hoffmann, and et~al., ``Stanley: The robot that won
  the darpa grand challenge,'' \emph{Journal of Field Robotics}, vol.~23,
  no.~9, p. 661–692, 2006. [Online]. Available:
  \url{https://dx.doi.org/10.1002/rob.20147}
\BIBentrySTDinterwordspacing

\bibitem{montemerlo2008}
\BIBentryALTinterwordspacing
M.~Montemerlo, J.~Becker, S.~Bhat, H.~Dahlkamp, D.~Dolgov, S.~Ettinger,
  D.~Haehnel, T.~Hilden, G.~Hoffmann, B.~Huhnke, and et~al., ``Junior: The
  stanford entry in the urban challenge,'' \emph{Journal of Field Robotics},
  vol.~25, no.~9, p. 569–597, 2008. [Online]. Available:
  \url{https://dx.doi.org/10.1002/rob.20258}
\BIBentrySTDinterwordspacing

\bibitem{Zablocki2021}
\BIBentryALTinterwordspacing
{\'{E}}.~Zablocki, H.~Ben-Younes, P.~P{\'{e}}rez, and M.~Cord,
  ``{Explainability of vision-based autonomous driving systems: Review and
  challenges},'' 2021. [Online]. Available:
  \url{http://arxiv.org/abs/2101.05307}
\BIBentrySTDinterwordspacing

\bibitem{Xiao2020}
Y.~Xiao, F.~Codevilla, A.~Gurram, O.~Urfalioglu, and A.~M. Lopez, ``{Multimodal
  End-to-End Autonomous Driving},'' \emph{IEEE Transactions on Intelligent
  Transportation Systems}, no. 201808390010, pp. 1--11, 2020.

\bibitem{chen2021}
J.~Chen, S.~E. Li, and M.~Tomizuka, ``Interpretable end-to-end urban autonomous
  driving with latent deep reinforcement learning,'' \emph{IEEE Transactions on
  Intelligent Transportation Systems}, pp. 1--11, 2021.

\bibitem{Behl2020}
A.~Behl, K.~Chitta, A.~Prakash, E.~Ohn-Bar, and A.~Geiger, ``{Label efficient
  visual abstractions for autonomous driving},'' in \emph{IEEE International
  Conference on Intelligent Robots and Systems}, 2020, pp. 2338--2345.

\bibitem{Prakash2020ExploringDA}
A.~Prakash, A.~Behl, E.~Ohn-Bar, K.~Chitta, and A.~Geiger, ``Exploring data
  aggregation in policy learning for vision-based urban autonomous driving,''
  \emph{2020 IEEE/CVF Conference on Computer Vision and Pattern Recognition
  (CVPR)}, pp. 11\,760--11\,770, 2020.

\bibitem{pmlr-v100-chen20a}
\BIBentryALTinterwordspacing
D.~Chen, B.~Zhou, V.~Koltun, and P.~Kr\"ahenb\"uhl, ``Learning by cheating,''
  in \emph{Proceedings of the Conference on Robot Learning}, ser. Proceedings
  of Machine Learning Research, L.~P. Kaelbling, D.~Kragic, and K.~Sugiura,
  Eds., vol. 100.\hskip 1em plus 0.5em minus 0.4em\relax PMLR, 30 Oct--01 Nov
  2020, pp. 66--75. [Online]. Available:
  \url{https://proceedings.mlr.press/v100/chen20a.html}
\BIBentrySTDinterwordspacing

\bibitem{ross11a}
\BIBentryALTinterwordspacing
S.~Ross, G.~Gordon, and D.~Bagnell, ``A reduction of imitation learning and
  structured prediction to no-regret online learning,'' in \emph{Proceedings of
  the Fourteenth International Conference on Artificial Intelligence and
  Statistics}, ser. Proceedings of Machine Learning Research, G.~Gordon,
  D.~Dunson, and M.~Dudík, Eds., vol.~15.\hskip 1em plus 0.5em minus
  0.4em\relax Fort Lauderdale, FL, USA: PMLR, 11--13 Apr 2011, pp. 627--635.
  [Online]. Available: \url{http://proceedings.mlr.press/v15/ross11a.html}
\BIBentrySTDinterwordspacing

\bibitem{Codevilla2019}
F.~Codevilla, E.~Santana, A.~Lopez, and A.~Gaidon, ``{Exploring the limitations
  of behavior cloning for autonomous driving},'' \emph{Proceedings of the IEEE
  International Conference on Computer Vision}, vol. 2019-October, no. Cvc, pp.
  9328--9337, 2019.

\bibitem{mao2021}
J.~Mao, M.~Niu, C.~Jiang, H.~Liang, X.~Liang, Y.~Li, C.~Ye, W.~Zhang, Z.~Li,
  J.~Yu, \emph{et~al.}, ``One million scenes for autonomous driving: Once
  dataset,'' \emph{arXiv preprint arXiv:2106.11037}, 2021.

\bibitem{Haan2019}
P.~D. Haan, D.~Jayaraman, and S.~Levine, ``Causal confusion in imitation
  learning,'' in \emph{NeurIPS}, 2019.

\bibitem{CodevillaICRA2018}
F.~Codevilla, M.~Müller, A.~López, V.~Koltun, and A.~Dosovitskiy,
  ``End-to-end driving via conditional imitation learning,'' in \emph{2018 IEEE
  International Conference on Robotics and Automation (ICRA)}, 2018, pp.
  4693--4700.

\bibitem{Hawke2020a}
J.~Hawke, R.~Shen, C.~Gurau, S.~Sharma, D.~Reda, N.~Nikolov, P.~Mazur,
  S.~Micklethwaite, N.~Griffiths, A.~Shah, and A.~Kndall, ``{Urban Driving with
  Conditional Imitation Learning},'' \emph{Proceedings - IEEE International
  Conference on Robotics and Automation}, pp. 251--257, 2020.

\bibitem{Huang2020}
Z.~Q. Huang, Z.~W. Qu, J.~Zhang, Y.~X. Zhang, and R.~Tian, ``{End-to-End
  Autonomous Driving Decision Based on Deep Reinforcement Learning},''
  \emph{Tien Tzu Hsueh Pao/Acta Electronica Sinica}, vol.~48, no.~9, pp.
  1711--1719, 2020.

\bibitem{osinski2020simulation}
B.~Osi{\'n}ski, A.~Jakubowski, P.~Ziecina, P.~Mi{\l}o{\'s}, C.~Galias,
  S.~Homoceanu, and H.~Michalewski, ``Simulation-based reinforcement learning
  for real-world autonomous driving,'' in \emph{2020 IEEE International
  Conference on Robotics and Automation (ICRA)}.\hskip 1em plus 0.5em minus
  0.4em\relax IEEE, 2020, pp. 6411--6418.

\bibitem{Liang_2018_ECCV}
X.~Liang, T.~Wang, L.~Yang, and E.~Xing, ``Cirl: Controllable imitative
  reinforcement learning for vision-based self-driving,'' in \emph{The European
  Conference on Computer Vision (ECCV)}, September 2018.

\bibitem{sutton1998introduction}
R.~S. Sutton, A.~G. Barto, \emph{et~al.}, \emph{Introduction to reinforcement
  learning}.\hskip 1em plus 0.5em minus 0.4em\relax MIT press Cambridge, 1998,
  vol. 135.

\bibitem{Pan2017b}
X.~Pan, Y.~You, Z.~Wang, and C.~Lu, ``{Virtual to real reinforcement learning
  for autonomous driving},'' \emph{British Machine Vision Conference 2017, BMVC
  2017}, 2017.

\bibitem{bdd100k}
F.~Yu, H.~Chen, X.~Wang, W.~Xian, Y.~Chen, F.~Liu, V.~Madhavan, and T.~Darrell,
  ``Bdd100k: A diverse driving dataset for heterogeneous multitask learning,''
  in \emph{IEEE/CVF Conference on Computer Vision and Pattern Recognition
  (CVPR)}, June 2020.

\bibitem{chen2017rethinking}
L.-C. Chen, G.~Papandreou, F.~Schroff, and H.~Adam, ``Rethinking atrous
  convolution for semantic image segmentation,'' \emph{arXiv preprint
  arXiv:1706.05587}, 2017.

\bibitem{Chen2018DeepLab}
L.-C. Chen, G.~Papandreou, I.~Kokkinos, K.~Murphy, and A.~L. Yuille, ``Deeplab:
  Semantic image segmentation with deep convolutional nets, atrous convolution,
  and fully connected crfs,'' \emph{IEEE Transactions on Pattern Analysis and
  Machine Intelligence}, vol.~40, no.~4, pp. 834--848, 2018.

\bibitem{tan2019efficientnet}
M.~Tan and Q.~Le, ``Efficientnet: Rethinking model scaling for convolutional
  neural networks,'' in \emph{International Conference on Machine
  Learning}.\hskip 1em plus 0.5em minus 0.4em\relax PMLR, 2019, pp. 6105--6114.

\bibitem{ronneberger2015u}
O.~Ronneberger, P.~Fischer, and T.~Brox, ``U-net: Convolutional networks for
  biomedical image segmentation,'' in \emph{International Conference on Medical
  image computing and computer-assisted intervention}.\hskip 1em plus 0.5em
  minus 0.4em\relax Springer, 2015, pp. 234--241.

\bibitem{espeholt2018impala}
L.~Espeholt, H.~Soyer, R.~Munos, K.~Simonyan, V.~Mnih, T.~Ward, Y.~Doron,
  V.~Firoiu, T.~Harley, I.~Dunning, \emph{et~al.}, ``Impala: Scalable
  distributed deep-rl with importance weighted actor-learner architectures,''
  in \emph{International Conference on Machine Learning}.\hskip 1em plus 0.5em
  minus 0.4em\relax PMLR, 2018, pp. 1407--1416.

\bibitem{horgan2018distributed}
D.~Horgan, J.~Quan, D.~Budden, G.~Barth-Maron, M.~Hessel, H.~Van~Hasselt, and
  D.~Silver, ``Distributed prioritized experience replay,'' \emph{arXiv
  preprint arXiv:1803.00933}, 2018.

\bibitem{espeholt2019seed}
L.~Espeholt, R.~Marinier, P.~Stanczyk, K.~Wang, and M.~Michalski, ``Seed rl:
  Scalable and efficient deep-rl with accelerated central inference,''
  \emph{arXiv preprint arXiv:1910.06591}, 2019.

\bibitem{cobbe2019quantifying}
K.~Cobbe, O.~Klimov, C.~Hesse, T.~Kim, and J.~Schulman, ``Quantifying
  generalization in reinforcement learning,'' in \emph{International Conference
  on Machine Learning}.\hskip 1em plus 0.5em minus 0.4em\relax PMLR, 2019, pp.
  1282--1289.

\bibitem{fujimoto2018addressing}
S.~Fujimoto, H.~Hoof, and D.~Meger, ``Addressing function approximation error
  in actor-critic methods,'' in \emph{International conference on machine
  learning}.\hskip 1em plus 0.5em minus 0.4em\relax PMLR, 2018, pp. 1587--1596.

\bibitem{haarnoja18b}
\BIBentryALTinterwordspacing
T.~Haarnoja, A.~Zhou, P.~Abbeel, and S.~Levine, ``Soft actor-critic: Off-policy
  maximum entropy deep reinforcement learning with a stochastic actor,'' in
  \emph{Proceedings of the 35th International Conference on Machine Learning},
  ser. Proceedings of Machine Learning Research, J.~Dy and A.~Krause, Eds.,
  vol.~80.\hskip 1em plus 0.5em minus 0.4em\relax PMLR, 10--15 Jul 2018, pp.
  1861--1870. [Online]. Available:
  \url{http://proceedings.mlr.press/v80/haarnoja18b.html}
\BIBentrySTDinterwordspacing

\bibitem{tobin2017domain}
J.~Tobin, R.~Fong, A.~Ray, J.~Schneider, W.~Zaremba, and P.~Abbeel, ``Domain
  randomization for transferring deep neural networks from simulation to the
  real world,'' in \emph{2017 IEEE/RSJ international conference on intelligent
  robots and systems (IROS)}.\hskip 1em plus 0.5em minus 0.4em\relax IEEE,
  2017, pp. 23--30.

\bibitem{laskin2020reinforcement}
M.~Laskin, K.~Lee, A.~Stooke, L.~Pinto, P.~Abbeel, and A.~Srinivas,
  ``Reinforcement learning with augmented data,'' \emph{arXiv preprint
  arXiv:2004.14990}, 2020.

\bibitem{ros}
\BIBentryALTinterwordspacing
{Stanford Artificial Intelligence Laboratory et al.}, ``Robotic operating
  system.'' [Online]. Available: \url{https://www.ros.org}
\BIBentrySTDinterwordspacing

\bibitem{onnxruntime}
O.~R. developers, ``Onnx runtime,'' \url{https://onnxruntime.ai/}, 2021.

\bibitem{kendall2019}
A.~Kendall, J.~Hawke, D.~Janz, P.~Mazur, D.~Reda, J.-M. Allen, V.-D. Lam,
  A.~Bewley, and A.~Shah, ``Learning to drive in a day,'' in \emph{2019
  International Conference on Robotics and Automation (ICRA)}, 2019, pp.
  8248--8254.

\end{thebibliography}

\end{document}